\documentclass[sigplan,screen]{acmart}
\usepackage{bm}
\usepackage{arydshln}
\usepackage{multirow}

\AtBeginDocument{%
  \providecommand\BibTeX{{%
    \normalfont B\kern-0.5em{\scshape i\kern-0.25em b}\kern-0.8em\TeX}}}

\setcopyright{acmcopyright}
\copyrightyear{2018}
\acmYear{2018}
\acmDOI{10.1145/1122445.1122456}

\acmConference[Woodstock '18]{Woodstock '18: ACM Symposium on Neural
  Gaze Detection}{June 03--05, 2018}{Woodstock, NY}
\acmBooktitle{Woodstock '18: ACM Symposium on Neural Gaze Detection,
  June 03--05, 2018, Woodstock, NY}
\acmPrice{15.00}
\acmISBN{978-1-4503-XXXX-X/18/06}

\begin{document}

\title{Few-shot Learning with Global Relatedness Decoupled-Distillation}

\author{Yuan Zhou$^{1}$,
        Yanrong Guo,
        Shijie Hao,
        Richang Hong, Zhengjun Zha, Meng Wang}
\author{$^{1}$ 2018110971@mail.hfut.edu.cn}
\renewcommand{\shortauthors}{Zhou et al.}

\begin{abstract}
Despite the success that metric learning based approaches have achieved in few-shot learning, recent works reveal the ineffectiveness of their episodic training mode. In this paper, we point out two potential reasons for this problem: 1) the random episodic labels can only provide limited supervision information, while the relatedness information between the query and support samples is not fully exploited; 2) the meta-learner is usually constrained by the limited contextual information of the local episode. To overcome these problems, we propose a new Global Relatedness Decoupled-Distillation (GRDD) method using the global category knowledge and the Relatedness Decoupled-Distillation (RDD) strategy. Our GRDD learns new visual concepts quickly by imitating the habit of humans, i.e. learning from the deep knowledge distilled from the teacher. More specifically, we first train a global learner on the entire base subset using category labels as supervision to leverage the global context information of the categories. Then, the well-trained global learner is used to simulate the query-support relatedness in global dependencies. Finally, the distilled global query-support relatedness is explicitly used to train the meta-learner using the RDD strategy, with the goal of making the meta-learner more discriminative. The RDD strategy aims to decouple the dense query-support relatedness into the groups of sparse decoupled relatedness. Moreover, only the relatedness of a single support sample with other query samples is considered in each group. By distilling the sparse decoupled relatedness group by group, sharper relatedness can be effectively distilled to the meta-learner, thereby facilitating the learning of a discriminative meta-learner. We conduct extensive experiments on the miniImagenet and CIFAR-FS datasets, which show the state-of-the-art performance of our GRDD method.
\end{abstract}



\begin{teaserfigure}
\centering
\includegraphics[width=16cm]{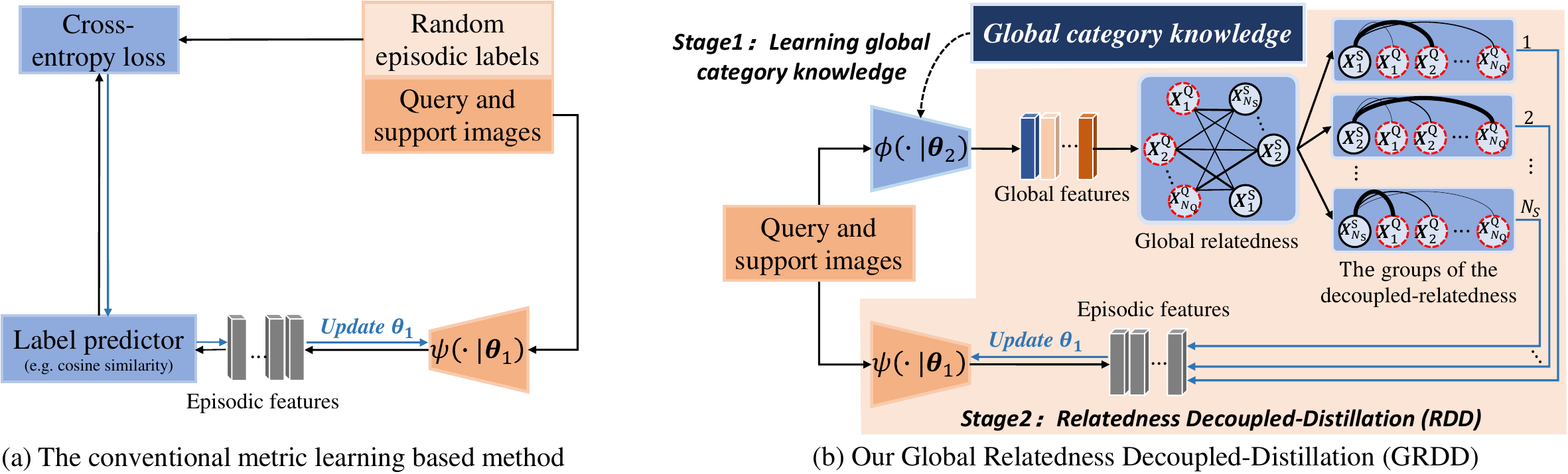}
\caption{The brief illustration of our Global Relatedness Decoupled-Distillation method (b) in training the meta-learner $\bm{\psi}(\cdot|\bm{\theta}_1)$, compared with the conventional metric learning based method (a). Of note, during the relatedness distillation, the well-trained global learner $\bm{\phi}(\cdot|\bm{\theta}_2)$ is frozen.}
\label{fig:0}
\end{teaserfigure}

\maketitle


\section{Introduction}
In recent years, deep learning has achieved impressive success in computer vision tasks such as image classification \cite{simonyan2014very,he2016deep}, object detection \cite{ren2016faster,redmon2016you} and semantic segmentation \cite{chen2017deeplab,fu2019dual}. However, it is well known that the deep-learning model tends to overfit for scarce training samples and performs far from satisfactory. In contrast, humans are still able to learn new visual concepts quickly in the data-scarce circumstances. This motivates the emergence of research for the few-shot learning (FSL) problem \cite{fe2003bayesian}, i.e., having the machine learning system quickly learn new visual concepts from only one or a few labelled training examples. \par

Intuitively, a straightforward solution to the overfitting problem is augmenting the target training dataset, e.g., by data synthesis \cite{schwartz2018delta,hariharan2017low} or large-scale weakly labelled or unlabelled datasets \cite{douze2018low,wu2018exploit}. However, the main problem of the data augmentation based approach is that the augmentation policies need to be tailored for different datasets due to domain gaps \cite{wang2020generalizing}. As one of the most widely used methods, the metric learning based method has achieved promising performance in FSL while maintaining high flexibility. In general, it aims to train a meta-learner for learning the transferable feature embeddings from the known categories (i.e., the categories of the auxiliary base subset $\bm{D}_{base}$ with adequate training data). To bridge the gaps between the training and testing phases, the episodic meta-training is designed \cite{vinyals2016matching}. Therefore, for the target FSL task whose sample categories are unobserved, the meta-learner first encodes the query and support samples into the embedding domain. Then, the query samples are matched with the support sample categories with the highest similarity \cite{vinyals2016matching} or the lowest distance \cite{snell2017prototypical}. \par

Despite the success achieved by metric learning based methods, recent works \cite{dhillon2019baseline,tian2020rethinking} show that their episodic training mode is ineffective or even unnecessary. In this paper, we first point out two potential reasons for this phenomenon: 1) the random episodic labels can only offer limited supervision information, 2) while the meta-learner is generally constrained by the limited intra- and inter-categorical context dependencies of the local episode. These issues limit the model's capability in producing high-quality transferable feature embeddings, and thus suppress model performance (the analysis is provided in Table \ref{tb:2} and Figure \ref{fig:4} of Section 4.3). To overcome these problems, we propose a new metric learning based method, named as Global Relatedness Decoupled-Distillation (GRDD), which mimics the human habit of learning new concepts quickly, i.e., learning from deep knowledge distilled by the teacher. The differences between our GRDD and the previous typical metric learning based methods are shown in Figure \ref{fig:0}. In the previous metric learning based methods (e.g. \cite{vinyals2016matching,snell2017prototypical,sung2018learning}), the meta-learner learns from the randomly constructed episodic labels whose supervision information is limited. In contrast, our GRDD utilizes the global relatedness between the query and support samples to train the meta-learner, which is more informative  and thus makes the learned transferable embeddings more discriminative. \par

\begin{figure*}[t!]
\centering
\includegraphics[height=7.5cm]{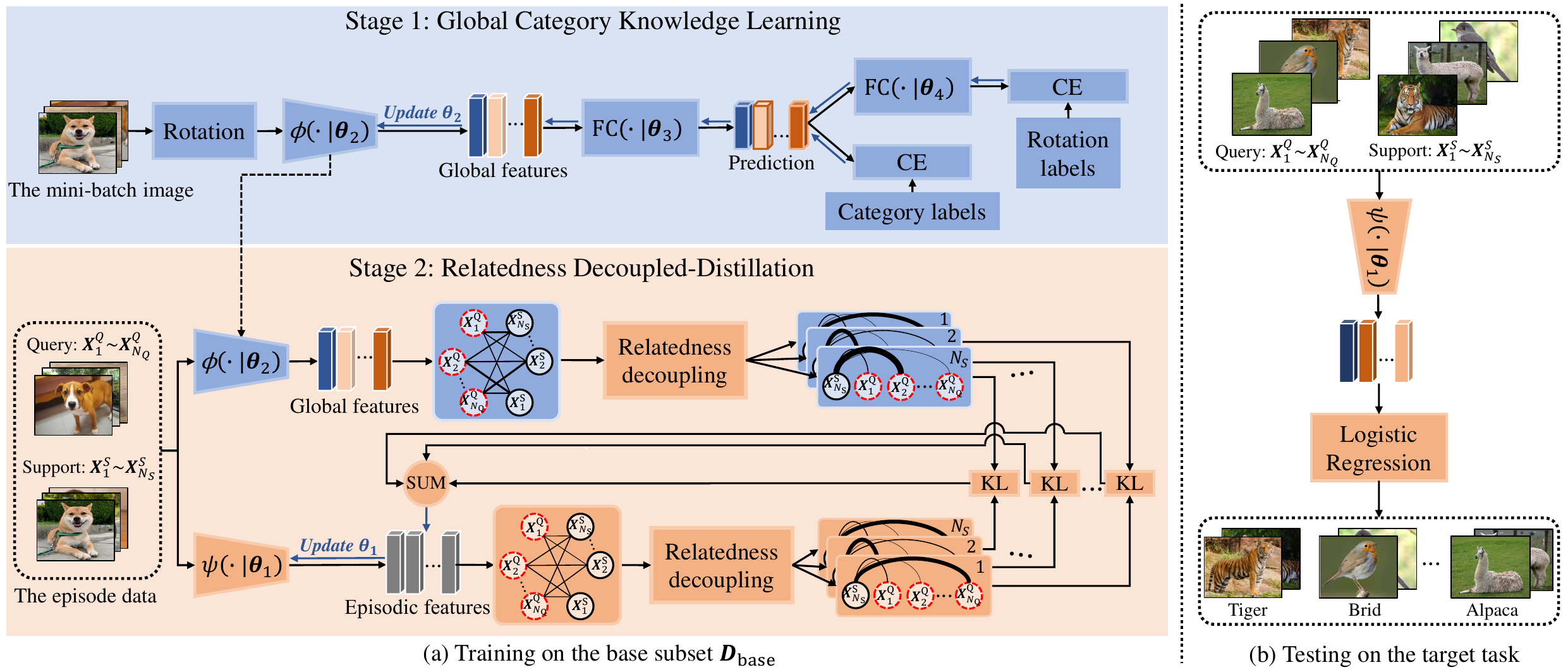}
\caption{The overview for our Global Relatedness Decoupled-Distillation (GRDD) method. In particular, `CE' stands for the cross-entropy loss and `KL' for the KL divergence, while $\bm{\psi}(\cdot|\bm{\theta}_1)$ and $\bm{\phi}(\cdot|\bm{\theta}_2)$ represent the meta-learner and the global-learner, respectively. Moreover, `FC' indicates the fully connected layer, and `SUM' denotes the summation operation.}
\label{fig:1}
\end{figure*}

As can be seen in Figure \ref{fig:1}, GRDD is designed in a two-stage training manner as dual-learners are used. In the first training stage, we train the global-learner $\bm{\phi}(\cdot|\bm{\theta}_2)$ on the entire base subset using category labels as supervision to fully exploit the global context dependencies of the categories. Then, in the second stage, the well-trained global-learner is used as a teacher to guide the episodic meta-training of the meta-learner $\bm{\psi}(\cdot|\bm{\theta}_1)$. Specifically, we first use the global-learner to simulate the global query-support relatedness for each episode, via leveraging the learned global category knowledge. Then, the global relatedness information is explicitly distilled to the meta-learner, which allows the meta-learner to know the samples' relatedness in the global context. To facilitate this process, we propose the Relatedness Decoupled-Distillation (RDD) strategy. It decouples the dense query-support relatedness into the groups of sparse decoupled relatedness. In particular, each group of decoupled relatedness only considers the relatedness of a single support sample with other query samples. On one hand, the sparser the relatedness is, the easier it can be distilled. On the other hand, decoupled relatedness is sharper in konwledge distillation, which is crucial in learning a discriminative meta-learner. To validate our method, extensive experiments are conducted on two public FSL datasets, i.e., miniImagenet \cite{vinyals2016matching} and CIFAR-FS \cite{bertinetto2018meta}, which firmly validates the effectiveness of our method. \par

All in all, the contribution of this paper can be summarized as follows:
\begin{itemize}
\item We point out the weaknesses of the current episodic training mode used in the metric learning based FSL methods, and propose a new Global Relatedness Decoupled-Distillation (GRDD) method to overcome these problems.
\item Instead of the random episodic labels, we propose to explicitly use the distilled global query-support relatedness to train the meta-learner, which makes the learned transferable feature embeddings more discriminative.
\item We introduce the Relatedness Decoupled-Distillation (RDD) strategy to facilitate the relatedness distillation. It decouples the entire query-support relatedness into the groups of sparse decoupled relatedness to make the relatedness information sharper and easier to be distilled.
\item On the miniImagenet and CIFAR-FS datasets, our proposed GRDD presents the state-of-the-art performance compared to other counterparts.
\end{itemize}

\section{Related work}
In this section, we briefly review the related FSL methods and introduce the differences between our proposed method and the most relevant approaches. \par

\textbf{Metric learning based method}. The metric learning based methods work in a learning-to-learn paradigm. It aims to train a meta-learner for learning high-quality transferable feature embeddings that can be well generalized to solve the target FSL tasks whose sample categories are unseen. Among the metric learning based methods, MatchNet \cite{vinyals2016matching} is a representative work. It
develops the episodic meta-training to bridge the gaps between the training and testing phases of FSL, using the random episodic labels as training supervision. Snell et al. \cite{snell2017prototypical} further develop MatchNet by introducing prototype representation so that query samples are categorized according to their Euclidean distances to the prototypes. Li et al. \cite{li2019few} propose to retrieve the global class representations by using the local features to categorize the query samples. Moreover, in \cite{li2020boosting}, they introduce the adaptive margin loss to improve the feature representation of the samples by further considering the semantic relation of the categories in Glove \cite{pennington2014glove}. Unlike the above methods that use random episodic labels as supervision \cite{vinyals2016matching,snell2017prototypical,li2019few,li2020boosting}, our GRDD explicitly uses the global query-support relatedness to train the meta-learning with the goal of making the meta-learner more discriminative. Moreover, compared to \cite{li2020boosting}, the distilled sample-wise relatedness is more fine-grained than the category-level relation in Glove \cite{pennington2014glove}, and thus more implicit information can be exploited, leading to accurate classification, as in Table \ref{tb:2}. Additionally, different from the works \cite{tian2020rethinking,rajasegaran2020self} simply resorting to the pretraining strategy, our GRDD aims to enhance the performance of the episodic training mode.\par

\textbf{External memory based method}. The external memory based methods are inspired by the recent success of Neural Turing Machine \cite{graves2014neural}. As a representative work, MAML \cite{finn2017model} proposes to design the memory module in a key-to-value paradigm. It first records the useful information of the support set into memory and then reads out the stored information to categorize the query samples. Ramalho et al. \cite{ramalho2018adaptive} boost the memory module by only memorizing the most unexpected information, thus suppressing memory redundancy. Kaiser et al. \cite{kaiser2017learning} design a long-term memory module suitable for solving the lifelong learning problems. It should be noted that memory-augmented models generally need to be fine-tuned on the support set of the target tasks in order to obtain sufficient useful information of the new categories. In contrast, our GRDD can be used to directly categorize the query samples  without the need for fine-tuning.\par

\textbf{Hallucination-based method}. The hallucination-based FSL methods can be divided into two sub-directions, i.e. hallucinations of new data \cite{hariharan2017low,wang2018low,zhang2019few} and hallucinations of classifier weights \cite{qiao2018few,gidaris2018dynamic,qi2018low}. Hariharan et al. \cite{hariharan2017low} propose a non-parametric data hallucination approach that hallucinates new support features for the novel unseen categories using the inter-category commonality. Wang et al. \cite{wang2018low} propose a hallucinator that synthesizes new images with different object poses or backgrounds by introducing random noise into the original image, while Zhang et al. \cite{zhang2019few} propose to hallucinate new data using the guidance of salient objects. In contrast to hallucinating new data, \cite{qiao2018few,gidaris2018dynamic,qi2018low} propose to hallucinate the classifier weights for the novel categories according to the feature activations of the support samples.\par

\textbf{Transductive vs. inductive method}. In traditional inductive FSL, each query sample is categorized independently. Transductive FSL, on the other hand, aims to categorize all query samples at once, or to consider the generated episodic tasks as a whole, thus leveraging information from both the support and the query sets. For example, Boudiaf et al. \cite{boudiaf2020information} propose to maximize the mutual information between the embedding features and the label prediction. Ziko et al. \cite{ziko2020laplacian} propose to impose an additional constraint on category inference, i.e., nearby samples should have the same consistent label assignments. In contrast, Liu et al. in \cite{liu2019learning} propose to propagate labels from labelled instances to unlabelled instances using the manifold structure of the data. As mentioned in \cite{ziko2020laplacian}, the transductive-based methods are usually more accurate than the inductive ones. Nevertheless, they face an unavoidable drawback, namely that the transductive model has to be retrained from scratch when new query samples or new episodic tasks appear. As for our GRDD, it is inductive and thus can be used to categorize new query samples or address new tasks directly once the training phase is complete. Last but not least, our GRDD can be easily integrated into the transductive approach, such as `GRDD-TIM' in Table \ref{tb:1}.

\section{Method}
In this section, we first present the preliminaries and then describe the proposed method in detail.

\subsection{Preliminary}
An FSL task consists of two subsets of data, commonly referred to as the support set $\bm{S}=\big\{(\bm{X}^S_i, \bar{y}^S_i)\big\}^{N_S-1}_{i=0}$ and query set $\bm{Q}=\big\{\bm{X}^Q_i\big\}^{N_Q-1}_{i=0}$. For the default `$C$-way $K$-shot' setting, the $N_S$ labelled support samples are prepared by randomly sampling $K$ labelled samples from each of the $C$ categories (i.e., $N_S=C\times K$), while the $N_Q$ unlabelled query samples are also randomly drawn from these $C$ categories. Note that the instances from the support and query sets are disjoint, i.e., $\bm{S}\cap\bm{Q}=\emptyset$. The ultimate goal of FSL is to categorize the query samples by exploiting the prior knowledge contained in the support set, as in Equation \ref{eq:0}:
\begin{equation}
y_i=\mathop{\arg\max}_{\tilde{y}_i \in \left\{1,..,C\right\}} P (\tilde{y}_i|\bm{X}_i^Q, \bm{S}).
\label{eq:0}
\end{equation}
In Equation \ref{eq:0}, $P (\tilde{y}_i|\bm{X}_i^Q, \bm{S})$ gives the probability that the query sample $\bm{X}_i^Q$ is classified as the label $\tilde{y}_i$ conditioned on the support set $\bm{S}$. To handle the FSL task, the metric learning based approach resorts to an auxiliary base subset $\bm{D}_{base}$. Note that the label spaces of the base subset $\bm{D}_{base}$ and the target FSL task are disjoint. As with current metric learning based methods, episodic meta-training is commonly used to bridge the gaps between the training and testing phases of FSL, using the randomly constructed episodic labels as training supervision, such as \cite{vinyals2016matching,snell2017prototypical,sung2018learning}. Obviously, the episodic training mode has advantages in training a meta-learner with high generalization. Nevertheless, the recent works (e.g., \cite{wang2020cooperative,tian2020rethinking}) reveal its ineffectiveness in training the FSL model. Therefore, this paper highlights two potential problems of the current episodic meta-training and proposes a new metric learning based method to alleviate these problems. \par


\subsection{Global Relatedness Decoupled-Distillation}
We propose a new metric learning based method, called as Global Relatedness Decoupled-Distillation (GRDD), which aims to imitate the human habit of learning novel concepts, i.e., learning from deep knowledge distilled by the teacher. In our GRDD method, two different learners are used, called global-learner $\bm{\phi}(\cdot|\bm{\theta}_2)$ and meta-learner $\bm{\psi}(\cdot|\bm{\theta}_1)$. Accordingly, as shown in Figure \ref{fig:1}, GRDD is designed in a two-stage training manner. In the first stage, the global-learner $\bm{\phi}(\cdot|\bm{\theta}_2)$ is trained on the entire base subset $\bm{D}_{base}$, using the category labels as training supervision. In this way, category knowledge can be exploited in the global contextual dependencies. In the second training stage, we then use the global query-support relatedness distilled from the global-learner $\bm{\phi}(\cdot|\bm{\theta}_2)$ to train the meta-learner $\bm{\psi}(\cdot|\bm{\theta}_1)$ based on the episodic training mode. To facilitate the relatedness learning, we propose the Relatedness Decoupled-Distillation (RDD) strategy in our GRDD, which decouples the dense query-support relatedness into the groups of sparse decoupled relatedness, making the relatedness sharper and easier to distill. Sections 3.2.1 and 3.2.2 present these two training stages in detail.

\subsubsection{Global Category Knowledge Learning}
To fully exploit the global context dependencies of the categories, we train the global-learner $\bm{\phi}(\cdot|\bm{\theta}_2)$ on the entire base subset $\bm{D}_{base}$ at the first training stage, using the category labels as supervision. In this process, we use the well-known mini-batch training strategy for fast model convergence. We also employ the data augmentation strategy  \cite{rajasegaran2020self}, which augments the input mini-batch images $\big\{ \bm{X}_i\big\}_{i=0}^{N_b-1}$ by rotating them to $90^\circ$, $180^\circ$, and $270^\circ$, respectively, and obtains $\big\{ \bm{X}_i^{R}\big\}_{i=0,R=0}^{N_b-1,270}$. Accordingly, the one-hot rotation labels are constructed, i.e., $\big\{\bar{r}^{R}_i\big\}^{N_b-1,270}_{i=0,R=0}$. Note that in the following sections $R=0$, $90$ , $180$ or $270$ is used when there are no special statements. As shown in Figure \ref{fig:1}, we first send the augmented mini-batch data to the global-learner $\bm{\phi}(\cdot|\bm{\theta}_2)$ and use the global-learner to extract their high-level feature representations $\big\{\bm{h}_i^{R}\big\}_{i=0,R=0}^{N_b-1,270}$ as in Equation \ref{eq:1}:

\begin{equation}
\bm{h}_i^{R} = \bm{\phi}(\bm{X}_i^{R}|\bm{\theta}_2)
\label{eq:1}
\end{equation}

\noindent where $\bm{h}_i^{R}$ is the features of $\bm{X}_i^{R}$, while $\bm{\theta}_2$ gives the learnable parameters of the global-learner. Then, a fully connected layer $FC(\cdot|\bm{\theta}_3)$ is applied to the features $\bm{h}_i^{R}$ to predict their categories, as in Equation \ref{eq:2}:

\begin{equation}
\bm{p}_i^{R} = FC(\bm{h}_i^{R}|\bm{\theta}_3).
\label{eq:2}
\end{equation}
In Equation \ref{eq:2}, $\bm{p}_i^{R}$ denotes the one-hot category prediction of $\bm{h}_i^{R}$, while $\bm{\theta}_3$ indicates the learnable parameters of the fully connected layer.\par

After that, another fully connected layer $FC(\cdot|\bm{\theta}_4)$ is applied to  $\bm{p}_i^{R}$, aiming to infer the label of rotation angle, as described in Equation \ref{eq:2_1}:

\begin{equation}
\bm{r}_i^{R} = FC(\bm{p}_i^{R}|\bm{\theta}_4).
\label{eq:2_1}
\end{equation}

Finally, the groundtruth category labels $\big\{\bar{y}_i\big\}_{i=0}^{N_b-1}$ and the rotation labels $\big\{\bar{r}^{R}_i\big\}^{N_b-1,270}_{i=0,R=0}$ are used to jointly optimize the entire network, as shown below:

\begin{equation}
\bm{\theta}_2^{\prime}=\bm{\theta}_2-lr_1*\frac{\partial{(\mathcal{L}_{c}+\mathcal{L}_{a})}}{\partial{\bm{\theta}_2}}
\label{eq:3}
\end{equation}
\begin{equation}
\bm{\theta}_3^{\prime}=\bm{\theta}_3-lr_1*\frac{\partial{(\mathcal{L}_{c}+\mathcal{L}_{a})}}{\partial{\bm{\theta}_3}}
\label{eq:4}
\end{equation}
\begin{equation}
\bm{\theta}_4^{\prime}=\bm{\theta}_4-lr_1*\frac{\partial{(\mathcal{L}_{c}+\mathcal{L}_{a})}}{\partial{\bm{\theta}_4}}
\label{eq:5}
\end{equation}
where
\begin{equation}
\mathcal{L}_{c}=\frac{1}{N_b*4}\sum^{N_b-1}_{i=0}\sum^{270}_{R=0}\sum^{C_{base}-1}_{j=0}one\_hot(\bar{y}_{i})_{j}*\log(\bm{p}_{i}^{R})_{j}
\label{eq:6}
\end{equation}
and
\begin{equation}
\mathcal{L}_{r}=\frac{1}{N_b*4}\sum^{N_b-1}_{i=0}\sum^{270}_{R=0}\sum^{3}_{j=0}one\_hot(\bar{r}^{R}_{i})_{j}*\log(\bm{r}_{i}^R)_{j}.
\label{eq:7}
\end{equation}
Note that the learning rate $lr_1$ is initialized as $5e^{-2}$ in Equation \ref{eq:3}, \ref{eq:4} and \ref{eq:5}, and decays in the `poly' manner. $one\_hot(\cdot)$ indicates the operation for one-hot encoding.\par

The advantages of learning global category knowledge are twofold. On one hand, the above training strategy is more global than the episodic training mode, and thus category knowledge can be learned in the more global contextual dependencies. On the other hand, the global knowledge learned by the global-learner $\bm{\phi}(\cdot|\bm{\theta}_2)$ is more informative than the random episodic labels, which can be used to better guide the training of the meta-learner $\bm{\psi}(\cdot|\bm{\theta}_1)$. In this way, the previously mentioned weaknesses of episodic meta-training can be relieved. In Section 3.2.2, we will elaborate the details of distiling the learned global knowledge to train the meta-learner $\bm{\psi}(\cdot|\bm{\theta}_1)$.\par

\subsubsection{Relatedness Decoupled-Distillation}
Considering that the random episodic labels can only provide limited supervision information, we therefore propose to use the global-learner $\bm{\phi}(\cdot|\bm{\theta}_2)$ for simulating the relatedness between the query and support samples in the global context dependencies of the categories, which is then used to explicitly train the meta-learner $\bm{\psi}(\cdot|\bm{\theta}_1)$. To facility the learning of relatedness, our GRDD method introduces the Relatedness Decoupled-Distillation (RDD) strategy.


More specifically, for each episodic data $\big\{\bm{X}^{S}_i,\bm{X}^{Q}_j\big\}_{i=0,j=0}^{N_S-1,N_Q-1}$, we first use the global-learner $\bm{\phi}(\cdot|\bm{\theta}_2)$, to extract their high-level features $\big\{\bm{h}_i^{S},\bm{h}_j^{Q}\big\}_{i=0,j=0}^{N_S-1,N_Q-1 }$. Since the features are extracted based on the learned global category knowledge, they are referred to as global features in this paper. Based on these global features, we then extract the global relatedness information between the query and support samples (i.e., $\bm{R}^{g}\in \mathbb{R}^{N_S\times N_q}$) using Equation \ref{eq:8}:

\begin{equation}
\bm{R}^{g}_{ij}=\frac{\bm{h}^{S}_{i}*\bm{h}^{Q}_{j}}{||\bm{h}^{S}_{i}||*||\bm{h}^{Q}_{j}||}
\label{eq:8}
\end{equation}
where $\bm{R}^{g}_{ij}$ denotes the $(i,j)$ element of $\bm{R}^{g}$ which is the relation between the $i$-$th$ support sample and the $j$-$th$ query sample. \par

Then, the relatedness $\bm{R}^{g}$ is decoupled into the groups of sparse decoupled relatedness $[\bm{\omega}^{g}_0,...,\bm{\omega}^{g}_{N_S-1}]$ for knowledge distillation, as in Equation \ref{eq:9}:
\begin{equation}
\bm{\omega}^{g}_{i}=\bm{\rVert}^{N_q-1}_{j=0}\frac{exp(\bm{R}_{ij}/T)}{\sum_{j}exp(\bm{R}_{ij}/T)}
\label{eq:9}
\end{equation}
where $\bm{\rVert}$ denotes the concatenation operation, while $T$ is the temperature hyperparameter used to smooth the values of relatedness for knowledge distillation.

After that, we distill the decoupled relatedness $[\bm{\omega}^{g}_0,...,\bm{\omega}^{g}_{N_S-1}]$ to the meta-learner $\bm{\psi}(\cdot|\bm{\theta}_1)$ group by group. For better explanation, we use $\bm{R}^{e}$ to represent the query-support relatedness computed based on the features extracted from the meta-learner, while the decoupled relatedness computed based on $\bm{R}^{e}$ is denoted as $[\bm{\omega}^{e}_0,...,\bm{\omega}^{e}_{N_S-1}]$. We first use KL divergence $KL(\cdot, \cdot)$ to measure the deviation between each group of $[(\bm{\omega}^{g}_0,\bm{\omega}^{e}_0),...,(\bm{\omega}^{g}_{N_S-1},\bm{\omega}^{e}_{N_S-1})]$, and then the loss of KL deviation is summed up as in Equation \ref{eq:10}:
\begin{equation}
\mathcal{L}_{kl}=\sum^{N_S-1}_{i=0} KL(\bm{\omega}^{g}_i,\bm{\omega}^{e}_i).
\label{eq:10}
\end{equation}
In addition, a regularized term $\mathcal{L}_{rt}$ is used to regularize the relatedness distillation, which constrains that samples with the same categories have higher relatedness:
\begin{equation}
\mathcal{L}_{rt}=\frac{1}{N_Q}\sum^{N_Q-1}_{i=0}\sum^{C-1}_{j=0}one\_hot(\bar{y}_{i}^{Q})_{j}*\log(\bm{\sigma}_{i})_{j}
\label{eq:11}
\end{equation}
where
\begin{equation}
\bm{\sigma}_i=\sum^{N_S-1}_{j=0} (\bm{R}^{e})^{T}_{ij}*one\_hot(\bar{y}_j^{S}).
\label{eq:12}
\end{equation}

\begin{table*}[t]
\footnotesize
\begin{center}
\caption{The accuracy comparison between our proposed GRDD and the related state-of-the-art approaches on  miniImagenet and CIFAR-FS datasets, with $95 \%$ confidence interval. It is noteworthy that the methods marked with `$^{\bm{\ddagger}}$' are based on transductive learning, while the remaining methods are inductive. Moreover, `Arch.' denotes the network architecture, while `n/a' indicates the unavailable results in original papers.}
\setlength{\tabcolsep}{3mm}{
\begin{tabular}{ccccccc}
\toprule[1pt]
 & & & \multicolumn{2}{c}{\textbf{miniImagenet, 5-way}} & \multicolumn{2}{c}{\textbf{CIFAR-FS, 5-way}}\\
\cline{4-5}\cline{6-7}
\multicolumn{1}{c}{\textbf{Method}} & \textbf{Reference} & \textbf{Arch.} & \multicolumn{1}{c}{1-shot} & \multicolumn{1}{c}{5-shot}& \multicolumn{1}{c}{1-shot}&\multicolumn{1}{c}{5-shot}\\
\hline
MatchNet \cite{vinyals2016matching} & NeurIPS' 16 & ConvNet-4 & $ 43.7\pm0.8$ & $55.3\pm0.7$ & $n/a$ & $n/a$ \\
MAML \cite{finn2017model}& ICML' 17 & ConvNet-4 & $48.7\pm1.8$  & $63.1\pm0.9$ & $58.9\pm1.9$  & $71.5\pm1.0$\\
ProtoNet \cite{snell2017prototypical}& NeurIPS' 17 & ConvNet-4 & $49.4\pm0.8$ & $68.2\pm0.7$ & $55.5\pm0.7$  & $72.0\pm0.6$ \\
DFS \cite{gidaris2018dynamic}& ICCV' 18  & ConvNet-4 & $ 56.2\pm0.9$ & $73.0\pm0.6$ & $n/a$ & $n/a$ \\
RelationNet \cite{sung2018learning}& CVPR' 18 & ConvNet-4 & $ 50.4\pm0.8$ & $65.3\pm0.7$ & $55.0\pm1.0$ & $69.3\pm0.8$\\
IMP \cite{allen2019infinite}& ICML' 19 & ConvNet-4 & $43.6\pm0.8$ & $55.3\pm0.7$ & $n/a$ & $n/a$ \\
TAML \cite{jamal2019task}& CVPR' 19 & ConvNet-4 & $51.8\pm1.9$ & $66.0\pm0.9$ & $n/a$ & $n/a$ \\
SAML \cite{hao2019collect}& ICCV' 19 & ConvNet-4 & $52.2\pm n/a$ & $66.5\pm n/a$ & $n/a$ & $n/a$ \\
GCR \cite{li2019few} & ICCV' 19 & ConvNet-4 & $53.2\pm0.8$ & $ 72.3\pm0.6$ & $n/a$ & $n/a$ \\
KTN \cite{peng2019few}& ICCV' 19 & ConvNet-4 & $54.6\pm0.8$ & $71.2\pm0.7$ & $n/a$ & $n/a$ \\
PARN \cite{wu2019parn} & ICCV' 19 & ConvNet-4 & $55.2\pm0.8$ & $71.6\pm0.7$ & $n/a$ & $n/a$ \\
R2D2 \cite{bertinetto2018meta} & ICLR' 19 & ConvNet-4 & $51.2\pm0.6$ & $68.8\pm0.1$ & $65.3\pm0.2$ & $79.4\pm0.1$\\
DC \cite{yang2021free} & ICLR' 21 & ConvNet-4 & $54.6\pm 0.6$ & $n/a$ & $n/a$ & $n/a$ \\
\cdashline{1-7}[0.8pt/2pt]

Our GRDD & \multirow{1}{*}{-} & ConvNet-4 & $\bm{58.9\pm 0.8}$ & $\bm{77.1\pm 0.6}$ & $\bm{69.3\pm 0.9}$ & $\bm{84.7\pm 0.6}$ \\

\hline
SNAIL \cite{mishra2018simple}& ICLR' 18 & ResNet-12 & $ 55.7\pm1.0$ & $68.9\pm0.9$  & $n/a$ & $n/a$ \\
AdaResNet \cite{munkhdalai2018rapid}& ICML' 18 & ResNet-12 & $ 56.9\pm0.6$ & $71.9\pm0.6$ & $n/a$ & $n/a$ \\
TADAM \cite{oreshkin2018tadam}& NeurIPS' 18 & ResNet-12 & $58.5\pm0.3$ & $76.7\pm0.3$ & $n/a$ & $n/a$ \\
Shot-Free \cite{ravichandran2019few}& ICCV' 19 & ResNet-12 & $ 59.0\pm n/a$ & $77.6\pm n/a$ & $69.2\pm n/a$ & $ 84.7\pm n/a$\\
TEWAM \cite{qiao2019transductive}& ICCV' 19 & ResNet-12 & $60.1\pm n/a$ & $75.9\pm n/a$ & $70.4\pm n/a$ & $81.3\pm n/a$ \\
MTL \cite{sun2019meta}& CVPR' 19 & ResNet-12 & $61.2\pm 1.8$ & $75.5\pm 0.8$ & $n/a$ & $n/a$ \\
VFSL \cite{schonfeld2019generalized}& CVPR' 19 & ResNet-12 & $61.2\pm 0.3$ & $77.7\pm 0.2$ & $n/a$ & $n/a$ \\
MetaOptNet \cite{lee2019meta}& CVPR' 19 & ResNet-12 & $ 62.6\pm 0.6$ & $78.6\pm 0.5$ & $72.6\pm0.7$ & $84.3\pm0.5$  \\
TRAML \cite{li2020boosting} & CVPR' 20 & ResNet-12 & $67.1\pm 0.5$ & $79.5\pm 0.6$ & $n/a$ & $n/a$ \\
DSN-MR & CVPR' 20 & ResNet-12 & $67.4 \pm 0.8$ & $82.9\pm 0.6$ &  $75.6\pm 0.9$ & $86.2\pm 0.6$\\
CBM \cite{wang2020cooperative} & MM' 20 & ResNet-12 & $64.8\pm 0.5$ & $80.5\pm 0.3$ & $n/a$ & $n/a$  \\
RFS \cite{tian2020rethinking}& Arxiv' 20 & ResNet-12 & $64.8\pm 0.6$ & $82.1\pm 0.4$ & $73.9\pm0.8$ & $86.9\pm0.5$ \\
SKD \cite{rajasegaran2020self} & Arxiv' 20 & ResNet-12 & $67.0\pm 0.9$ & $83.5\pm 0.5$ & $76.9\pm0.9$ & $88.9\pm0.6$  \\

\cdashline{1-7}[0.8pt/2pt]
Our GRDD & - & ResNet-12 & $\bm{67.5\pm 0.8}$ & $\bm{84.3\pm 0.5}$ & $\bm{77.5\pm 0.9}$ & $\bm{89.1\pm 0.6}$ \\
\hline
TPN \cite{liu2019learning}$^{\bm{\ddagger}}$ & ICLR' 19 & ConvNet-4 & $55.5\pm 0.9$ & $69.9\pm 0.7$ & $n/a$  & $n/a$ \\
Feat \cite{ye2020few}$^{\bm{\ddagger}}$  & CVPR' 20 & ConvNet-4 & $57.0\pm 0.2$ & $72.9\pm 0.2$ & $n/a$ & $n/a$ \\
MRN \cite{he2020memory}$^{\bm{\ddagger}}$ & MM' 20 & ConvNet-4 & $57.8\pm 0.7$ & $71.1\pm 0.5$ & $n/a$ & $n/a$ \\

\cdashline{1-7}[0.8pt/2pt]
Our GRDD-TIM$^{\bm{\ddagger}}$ & - & ConvNet-4 & $\bm{65.7\pm 0.3}$ & $\bm{80.1\pm 0.2}$ & $\bm{79.9\pm 0.2}$ & $\bm{87.9\pm 0.2}$ \\

\hline

LaplacianShot \cite{ziko2020laplacian}$^{\bm{\ddagger}}$ & ICML' 20 & ResNet-18 & $72.1\pm 0.2$ & $82.3\pm 0.1$ & $n/a$ & $n/a$ \\
TIM \cite{boudiaf2020information}$^{\bm{\ddagger}}$ & NeurIPS' 20 & ResNet-18 & $73.9\pm 0.2$ & $85.0\pm 0.1$ & $n/a$ & $n/a$ \\
BD-CSPN \cite{liu2020prototype}$^{\bm{\ddagger}}$ & ECCV' 20 & WRN-28-10 & $70.3\pm 0.9$ & $81.9\pm 0.6$ & $n/a$ & $n/a$ \\

IFSL-SIB \cite{yue2020interventional}$^{\bm{\ddagger}}$ & NeurIPS' 20 & WRN-28-10 & $73.5\pm n/a$ & $83.2\pm n/a$ & $n/a$ & $n/a$ \\

\cdashline{1-7}[0.8pt/2pt]
Our GRDD-TIM$^{\bm{\ddagger}}$ & - & ResNet-12 & $\bm{75.8\pm0.2}$ & $\bm{87.3\pm0.1}$ & $\bm{85.4\pm 0.2}$  &  $\bm{91.1\pm 0.2}$ \\

\bottomrule[1pt]
\end{tabular}}
\label{tb:1}
\end{center}
\end{table*}

Finally, the meta-learner parameters are updated via the joint usage of $\mathcal{L}_{kl}$ and $\mathcal{L}_{rt}$, as shown in Equation \ref{eq:13}:
\begin{equation}
\bm{\theta}_1^{\prime} = \bm{\theta}_1 - lr_{2}*\frac{\partial{(\mathcal{L}_{kl}+\gamma*\mathcal{L}_{rt})}}{\partial{\bm{\theta}_1}}
\label{eq:13}
\end{equation}
where the hyperparameter $\gamma$ is to control the balance between $\mathcal{L}_{kl}$ and $\mathcal{L}_{rt}$. Note that during this process, the parameters of the well-trained global-learner $\bm{\phi}(\cdot|\bm{\theta}_2)$ are frozen. By using the proposed RDD strategy to train the meta-learner $\bm{\psi}(\cdot|\bm{\theta}_1)$, our method achieves competitive experimental performance compared to other counterparts, which is shown in the next section.

\begin{table}[t]
\footnotesize
\begin{center}
\caption{The demonstration for the weaknesses of the current episodic meta-training and the strength of our relatedness distillation method in FSL. `CL' or `EL'  denotes the experiments that use category labels or episodic labels as training supervision. `GR' represents the usage of our global relatedness. `Arch.' denotes the network architecture.}
\setlength{\tabcolsep}{2.5mm}{
\begin{tabular}{ccccc}
\toprule[1pt]
Supervision &  Arch. of $\bm{\phi}(\cdot|\bm{\theta}_2)$ & Arch. of $\bm{\psi}(\cdot|\bm{\theta}_1)$ & Acc (\%)  \\
\hline
CL & $no$  & ConvNet-4& $64.6\pm 0.9$  \\
CL+EL & $no$  & ConvNet-4&  $66.7\pm 0.9$ \\
CL+GR & ConvNet-4  & ConvNet-4& $67.3\pm 0.9$  \\
CL+GR & ResNet-12  & ConvNet-4& $69.3\pm 0.9$  \\
\cdashline{1-4}[0.8pt/2pt]
CL & $no$  & ResNet-12 & $74.9\pm 0.9$  \\
CL+EL & $no$  & ResNet-12 & $72.9\pm 0.9$ \\
CL+GR & ConvNet-4  & ResNet-12& $75.4\pm 0.9$  \\
CL+GR & ResNet-12  & ResNet-12& $77.5\pm 0.9$  \\
\bottomrule[1pt]
\end{tabular}}
\label{tb:2}
\end{center}
\end{table}

\section{Experiment}
With the aim of validating our proposed method, we conduct extensive experiments on two public FSL datasets, i.e., miniImagenet \cite{vinyals2016matching} and CIFAR-FS \cite{bertinetto2018meta}. In this section, we first introduce these datasets and the implementation details of the experiments. Then, we compare our GRDD in detail with the related state-of-the-art approaches.

\subsection{Dataset and Implementation Details}
\textbf{Dataset}. miniImagenet \cite{vinyals2016matching} and CIFAR-FS \cite{bertinetto2018meta} are the most commonly used FSL datasets. In particular, CIFAR-FS is derived from the CIFAR-100 \cite{krizhevsky2009learning} dataset, while miniImagenet is derived from the larger ILSVRC-12 \cite{russakovsky2015imagenet} dataset. Remarkably, these two datasets both contain 60000 images with 100 different semantic categories. But the  image resolutions of the datasets are different. Specifically, CIFAR-FS consists of $32\times 32$ images, while the images from miniImagenet have a resolution of $84\times 84$. Following previous works \cite{rajasegaran2020self,wang2020cooperative,he2020memory}, for these two datasets, the 100 categories are divided into 64, 16 and 20 for training, validation and testing, respectively.

\textbf{Implementation Details}. All our experiments are built on Pytorch\footnote{\url{https://pytorch.org/}}. Following \cite{wang2020cooperative, snell2017prototypical, bertinetto2018meta}, we respectively use ConvNet-4 \cite{vinyals2016matching} and ResNet-12 \cite{he2016deep} to implement the meta-learner. Note that the global-learner is implemented by ResNet-12  if there is no special declaration. For all experiments, we choose the Stochastic Gradient Descent (SGD) as the optimizer, of which the weight decay is empirically set to $5e^{-4}$. Under the two-stage training manner, different training strategies are applied to the different stages. In particular, for the first training stage, we adopt the `poly' learning rate, i.e., $lr_{1}=lr\_{init}\times (1-\frac{iter}{iter\_{total}})^{power}$, where $lr\_{init}$ is set to $1e^{-1}$ and $power$ is set to $0.9$. We also use the well-known mini-batch training strategy
for fast model convergence. Note that for all datasets, the batch size is set to 64 and the epoch is set to 90. However, in the second training stage, we set a smaller initial learning rate $lr_2$ and epoch, which are $1e^{-3}$ and $15$, respectively. Moreover, the learning rate decays by a factor of 0.1 for the last 5 epochs. For the hyperparameters, $\gamma$ and $T$ are respectively set to $0.2$ and $4$, which are validated in Section 4.3.

\subsection{Comparison with the state-of-the-art methods}
In this section, we compare our GRDD with related state-of-the-art approaches summarized in Table \ref{tb:1}. Note that for a fair comparison, the methods based on different network structures are compared accordingly. \par

\textbf{ConvNet-4}. In this part, we compare our GRDD with the methods implemented using ConvNet-4. As shown in Table \ref{tb:1}, our GRDD largely outperforms the compared methods on the  miniImagenet and CIFAR-FS datasets. For example, on the miniImagenet dataset, GRDD is more accurate than DC \cite{yang2021free} and R2D2 \cite{bertinetto2018meta}  by about $4 \%$ and $8 \%$, respectively. On the CIFAR-FS dataset, the accuracy of our GRDD is still higher than that of R2D2, about $4 \%$ higher on the 5-way 1-shot task, while $5 \%$ higher on the 5-way 5-shot task. \par

\textbf{ResNet-12}. In general, higher accuracy can be achieved by using larger models. Thus, by implementing GRDD with ResNet-12, GRDD consistently shows better performance than the ConvNet-4 version. As shown in Table \ref{tb:1}, our GRDD is obviously more accurate than most methods, such as SNAIL \cite{mishra2018simple} and VFSL \cite{schonfeld2019generalized}. Moreover, our GRDD is even better compared to the recent works RFS \cite{tian2020rethinking}, SKD \cite{rajasegaran2020self} and CBM \cite{wang2020cooperative}. For example, on the miniImagenet dataset, the accuracy of our method is  $0.5 \%$ and $0.8 \%$ higher than that of SKD on the 1-shot and 5-shot tasks, respectively. On the CIFAR-FS dataset, the accuracy of our GRDD is about $4 \%$ and $2 \%$ more accurate than RFS, respectively.

\textbf{Transductive learning}. Although our GRDD is proposed as an inductive approach, it can be easily integrated into the transductive learning approach. For example, we integrate our GRDD with TIM \cite{boudiaf2020information}, which is called `GRDD-TIM' in Table \ref{tb:1}. On the one hand, `GRDD-TIM' achieves a significant performance gain over the baseline TIM \cite{ziko2020laplacian}. On the other hand, it also achieves the state-of-the-art performance among the transductive counterparts even if our GRDD is implemented based on a smaller neural network ResNet-12. \par

The above experiments strongly firm the effectiveness and flexibility of our GRDD. It should be noted that the ablation study is conducted in Section 4.3 to further analyze our proposed method.

\begin{figure}[t!]
\centering
\includegraphics[height=4.2cm]{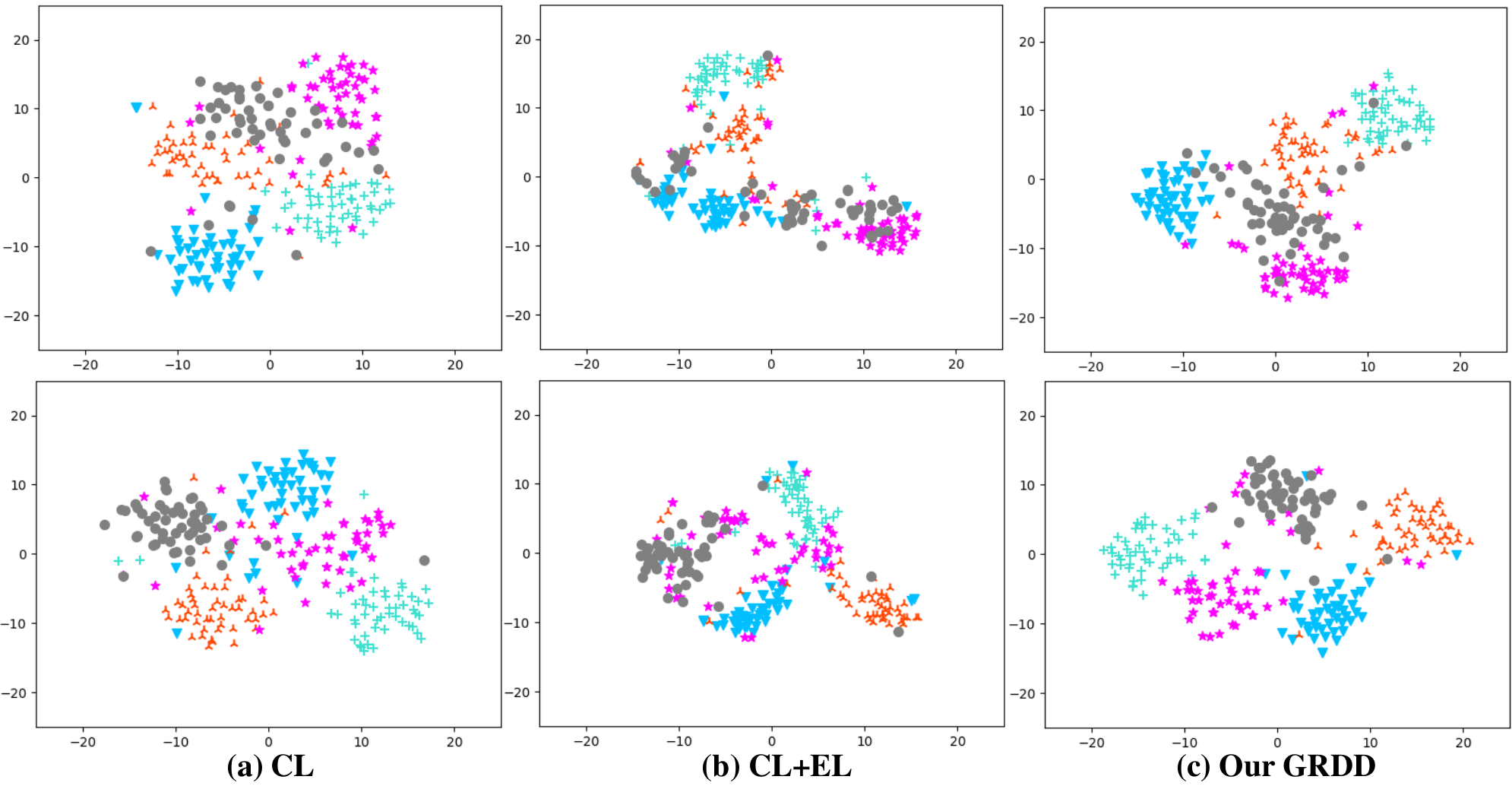}
\caption{The t-SNE visualization for the embeddings of the `CL', `CL+EL' and our GRDD method based on ResNet-12.}
\label{fig:4}
\end{figure}

\begin{figure}[t!]
\centering
\includegraphics[height=3.2cm]{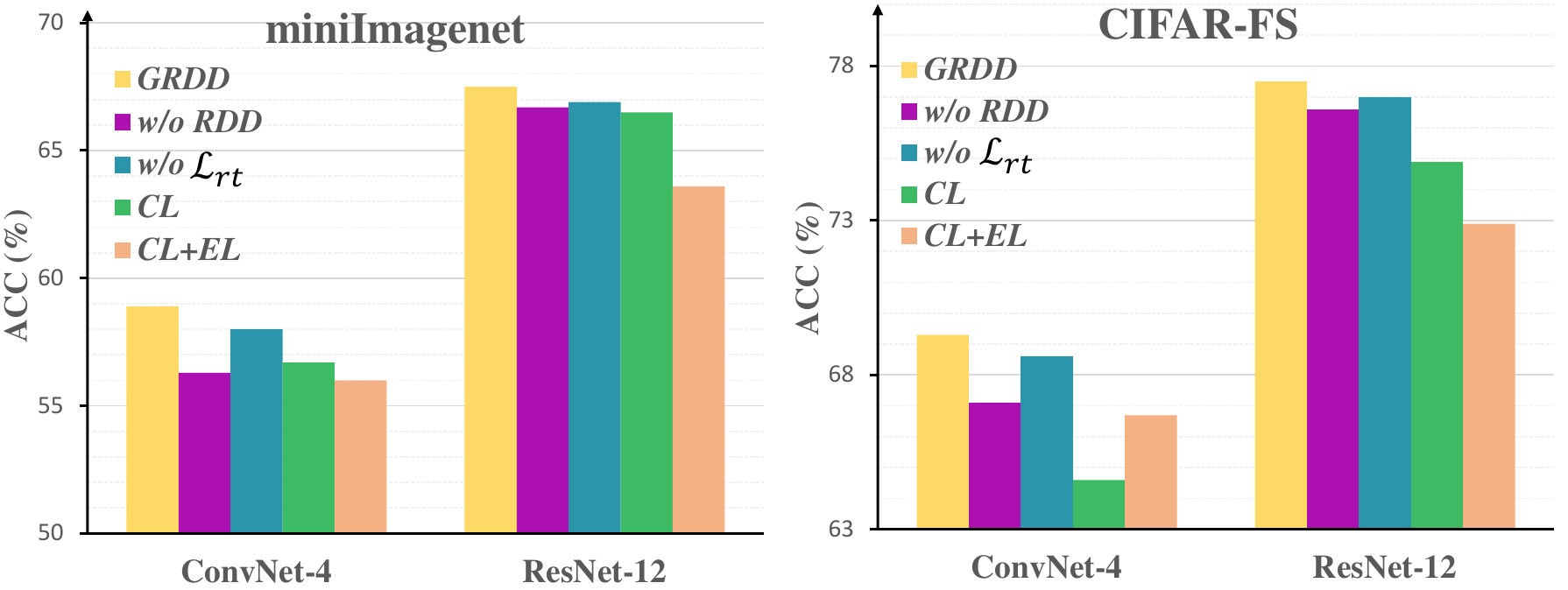}
\caption{The influence of each component of GRDD. `w/o RDD' indicates that GRDD is used without the Relatedness Decoupled-Distillation (RDD) strategy and the relatedness is distilled as a whole instead. `w/o $\mathcal{L}_{rt}$' indicates that GRDD is implemented without using the regularized term $\mathcal{L}_{rt}$.}
\label{fig:2}
\end{figure}

\begin{figure}[t!]
\centering
\includegraphics[height=5.2cm]{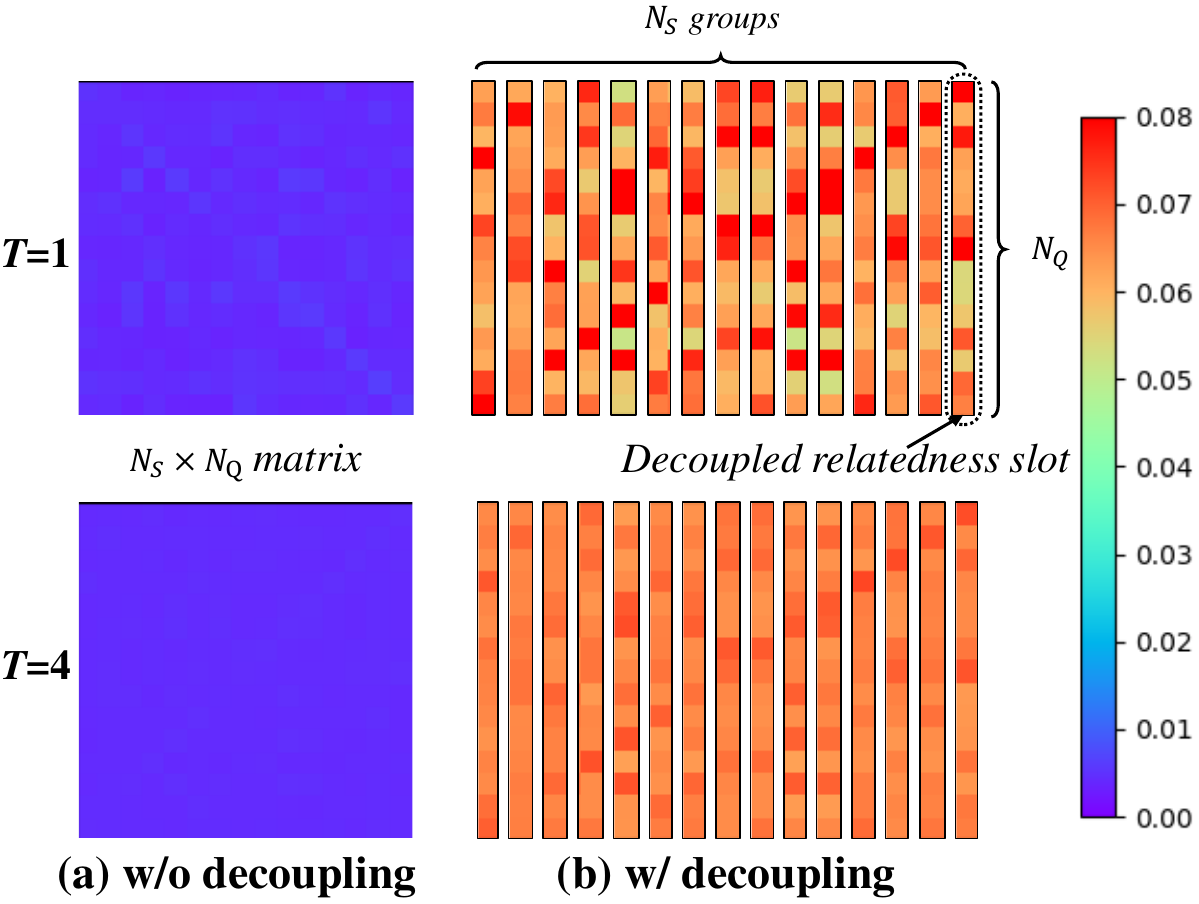}
\caption{The comparison between relatedness with (b) and without (a) relatedness decoupling in knowledge distillation, under different values of temperature $T$. Note that without decoupling, the relatedness is distilled as a whole matrix. While with decoupling, the relatedness is distilled corresponding to each decoupled relatedness slot.}
\label{fig:5}
\end{figure}

\begin{figure}[t!]
\centering
\includegraphics[height=3.5cm]{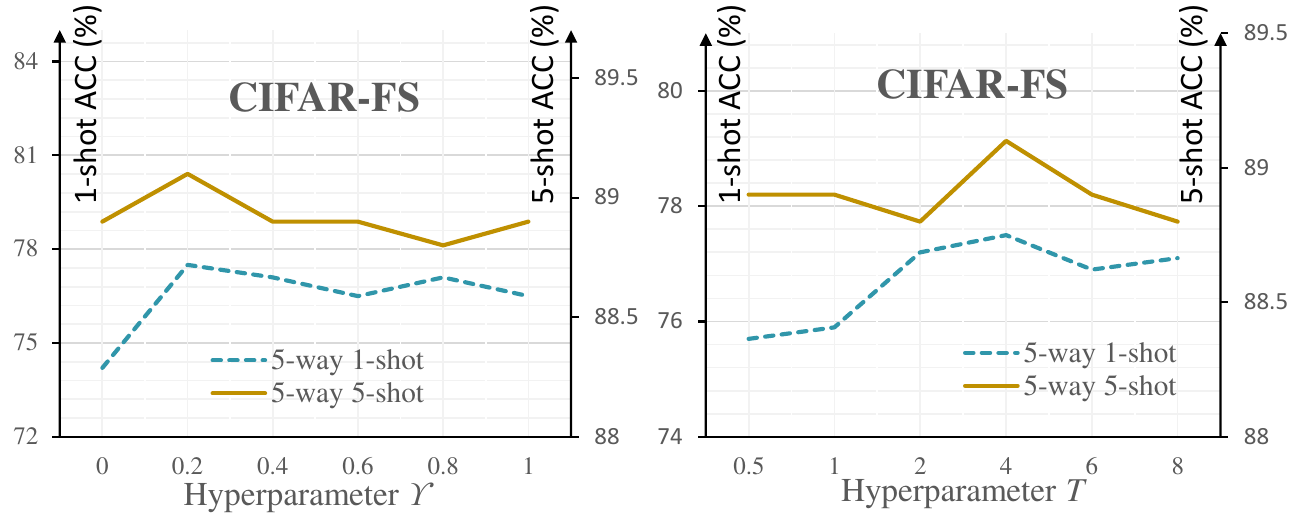}
\caption{Ablation study for the hyperparameters $\gamma$ and $T$. }
\label{fig:3}
\end{figure}

\subsection{Ablation study}
In this section, we conduct the ablation study for our work. We first analyze the weaknesses of the current episodic training mode. Then, we investigate the impact of each component in our GRDD. Finally, the settings of two vital haperparameters (i.e., $\gamma$ and $T$) are validated. For brevity, we note that `CL' denotes the methods pretrained on the Category Labels, while ``CL+EL'' indicates the methods that further finetrain the pretrained model using the Episodic Labels. Moreover, `CL+GR' denotes our GRDD that uses the Global Relatedness extracted from the category labels to train the meta-learner. \par

\textbf{Improvement over episodic training}. Recent works \cite{tian2020rethinking,wang2020cooperative} find that the episodic training mode in FSL is ineffective and unnecessary. Here, we give two potential reasons for this phenomenon, which are experimentally analyzed in this part. As shown in Table \ref{tb:2}, `CL+EL' does not always yield a performance gain over the baseline model `CL'. For example, on ResNet-12, accuracy actually decreases by about $2 \%$ when episodic labels are further used. This is because the episodic labels can only provide limited supervision and thus are unable to boost the quality of feature embeddings effectively. Instead, the learned global category knowledge may be destroyed by the local episodic meta-training, whose context is very limited. However, when more informative global relatedness is used in meta-training, `CL+GR' achieves significant improvement in all experiments. In addition, the more accurate the relatedness information is (i.e., extracted by a larger model), the higher the accuracy can be obtained. This proves the effectiveness of our GRDD, while the limited information of the episodic labels is the bottleneck in the episodic training mode. This conclusion is also consistent with the visualized analysis in Figure \ref{fig:4}, where we can see that the episodic labels make the embedding space more compact, but the boundary between different categories becomes blurred because of the limited guidance of supervision information. However, our relatedness information makes the embedding space more compact, meanwhile the category boundary becomes clearer and more discriminative.\par

\textbf{Influence of each component in GRDD.} As shown in Figure \ref{fig:2}, we first compare our GRDD with two degenerate versions `w$/$o $\mathcal{L}_{rt}$' and `w$/$o RDD'. The results in Figure \ref{fig:2} indicate: 1) using the RDD strategy is better than distilling the relatedness information as a whole matrix; 2) incorporating RDD with the regularized term $\mathcal{L}_{rt}$ is better than using RDD alone.  Moreover, our GRDD also shows consistently better performance than `CL' and `CL+EL'. Therefore, the effectiveness of the two key components of our GRDD can be verified. It is also worth noting that the visualization of the relatedness with and without decoupling in the knowledge distillation is demonstrated in Figure \ref{fig:5}, where the decoupled relatedness is more discriminative than the relatedness that is considered as a whole matrix.


\textbf{Hyperparameter settings}. Furthermore, the experiments in Figure \ref{fig:3} are conducted to validate the settings for two key hyperparameters in our GRDD, i.e., $\gamma$ and $T$. The results in Figure \ref{fig:3} show that the $\gamma=0.2$ and $T=4$ setting can yield better performance. \par

\section{Conclusion}
In this paper, we show that the bottleneck of the episodic training mode lies in the limited supervision information of episodic labels and the scarce category context. To alleviate these problems, we propose a new Global Relatedness Decoupled-Distillation (GRDD) method that explicitly uses the more informative global query-support relatedness to train the meta-learner, making it more discriminative. Moreover, the Relatedness Decoupled-Distillation (RDD) strategy is introduced to facilitate this procedure. RDD decouples the dense relatedness into the groups of sparse decoupled relatedness, making the relatedness sharper and easier to be distilled.  Extensive experiments on the miniImagenet and CIFAR-FS datasets validate the effectiveness of our method. In the future, we plan to apply our method in other FSL domains, such as open-set FSL and domain-shift FSL.
\clearpage
\bibliographystyle{ACM-Reference-Format}
\bibliography{sample-base}

\end{document}